\pdfoutput=1

\documentclass[11pt,a4paper]{article}

\usepackage{authblk}

\usepackage[hyperref]{acl2021}

\usepackage[hyphenbreaks]{breakurl}

\usepackage{times}
\usepackage{latexsym}
\usepackage{amssymb}

\usepackage{microtype}

\usepackage{comment}
\usepackage{censor}

\aclfinalcopy 
\def\aclpaperid{***} 
\StopCensoring

\title{
\censor{Handshakes AI Research}
at CASE 2021 Task 1: \\
Exploring different approaches for multilingual tasks
}

\author[]{Vivek Kalyan\protect\thanks{\hspace{0.5em}Equal contributions}}
\newcommand\CoAuthorMark{\footnotemark[\arabic{footnote}]}
\author[]{Paul Tan\protect\CoAuthorMark}
\author[]{Shaun Tan\protect\CoAuthorMark}
\author[]{Martin Andrews}

\affil[]{%
  Handshakes, Singapore\\
  \url{{first.last}@handshakes.com.sg}%
}

\date{}

\begin{document}
\maketitle
\begin{abstract}
The aim of the CASE 2021 Shared Task~1 \cite{Hurriyetoglu+21-task1} 
was to detect and classify socio-political and crisis event information 
at document, sentence, cross-sentence, and token levels in a multilingual setting, 
with each of these subtasks being evaluated separately in each test language. 
Our submission contained entries in all of the subtasks, 
and the scores obtained validated our research finding: 
That the multilingual aspect of the tasks should be embraced, 
so that modeling and training regimes use the multilingual nature of the
tasks to their mutual benefit, 
rather than trying to tackle the different languages separately.
Our code is available at 
\url{https://github.com/HandshakesByDC/case2021/}
\end{abstract}

\section{Introduction}


The CASE Shared Task 1 concerned news events that are in the scope of contentious politics and characterized by riots and social movements, denoted ``GLOCON Gold'' 
\cite{Hurriyetoglu+20b}. 
The aim of the shared task was to detect and classify socio-political and crisis event information at document, sentence, cross-sentence, and token levels in a multilingual setting:

\begin{itemize}
\item Subtask 1 : Document classification: Does a news article contain information about a past or ongoing event?
\item Subtask 2 : Sentence classification: Does a sentence contain information about a past or ongoing event?
\item Subtask 3 : Event sentence coreference identification: Which event sentences (from Subtask 2) are about the same event?
\item Subtask 4 : Event extraction: What is the event trigger and its arguments?
\end{itemize}

The detailed description of the subtasks can be found in \citet{Hurriyetoglu+19b} and \citet{10.1162/dint_a_00092}.

\section{Team Organisation}

In order to efficiently allocate resources,
separate, parallel research efforts were initially made towards each subtask, 
with periodic knowledge sharing taking place between subtasks.  

Data issues with Subtask~1 
(whereby, due to copyright reasons, 
a significant number of the news articles were severely truncated in the dataset provided),
our original approach to this subtask was abandoned, 
and the approach from Subtask~2 was quickly redeployed towards Subtask~1 
in the late stages of the Shared Task test phase 
%
- hence the ordering herein of system descriptions.


\section{Methods}

All subtask teams used off-the-shelf pre-trained models, 
and training was conducted only on the training data provided through the Shared Task 
(except as noted in Subtask~3, where some additional public data was used).


The key language models used for the subtasks were 
pre-trained models sourced from the Hugging Face library\footnote{\url{https://huggingface.co/models}}:

\begin{itemize}
\item DistilBERT, Multilingual (`m-distilBERT') \cite{Sanh2019DistilBERTAD}
\item BERT-Base, Multilingual Cased (`m-BERT') \cite{devlin-etal-2019-bert} 
\item `XLM-RoBERTa' (multilingually trained, \mbox{-base} version) \cite{conneau2020unsupervised}
\end{itemize}

For generating embeddings for sentences, 
and as part of the word-at-a-time translation technique used in Subtask~4, 
we used the following publicly available pre-trained models:

\begin{itemize}
\item `LASER' (Language-Agnostic SEntence Representations) \cite{artetxe2019massively} 
\item Language-agnostic BERT Sentence Embedding (`LaBSE') \cite{feng2020languageagnostic} 
\item Multilingual Universal Sentence Encoder (`M-USE') \cite{yang-etal-2020-multilingual}
\item Multilingual Unsupervised and Supervised Embeddings (`MUSE') \cite{lample2017unsupervised}
\end{itemize}

Due to the use of pre-trained models, 
the computational resources required no more than single-GPU workstations.

\section{Subtask System Descriptions}

\subsection{Subtask 2 - Sentence Classification}

\noindent\fbox{%
    \parbox{\linewidth}{%
Does a sentence contain information about a past (or ongoing) event, or not? (Binary classification)
    }%
}

\subsubsection{Experimental Approach}
The sentence classification subtask had a relatively 
high quantity of training data
with all test languages having corresponding training data.
Our approach was to find the best combined training dataset 
to train the largest multilingual model available.

To create internal classification baselines, 
we initially used a linear classifier over LASER embeddings and then progressed to m-distilBERT. 
Then, 
using the efficient pipeline created, 
we performed ablation tests to select the best training dataset across all models,
from among the training datasets that we constructed.

The remaining time was spent fine-tuning the largest multilingual model available, XLM-RoBERTa. 
Based on our experimental results, 
we decided to train a single model to generate the final submission on all languages.

\subsubsection{Model and Data Architecture}
Our final training dataset used the training data from all languages 
into a single combined dataset.
This dataset was split $80/20$ for training and internal validation sets. 

Our final model was a pre-trained XLM-RoBERTa model, 
fine-tuned on the article data from Subtask~1 and Subtask~2, 
with a `classification head' 
(i.e. a single linear layer on top of the pooled output from the transformer layers) 
trained on the Subtask~2-specific training data.
For the classification component, we selected the model that maximised validation $F_1$ scores, 
and our component scores are listed in Table \ref{table:subtask2}.

\subsubsection{Experimental Results}

We found that the best performing training dataset
was made by combining all 3 datasets provided in their original language
into a single all-encompassing dataset : The multilingual model benefiting
from seeing all of the data as one coherent set.

\begin{table}[h]

\begin{tabular}{lrrr}
\hline
Dataset & English & Spanish & Portuguese\\
\hline
Validation & 0.7610 & 0.6950 & 0.6670\\
Competition & 0.7750 & 0.8325 & 0.8506\\
Final Placing & 7/11 & 3/10 & 4/10\\
\hline
\end{tabular}
\caption{Averaged Model Performance for Subtask~2} 
\label{table:subtask2} 
\end{table}

Performance on Spanish and Portuguese showed good improvements by training on
all data instead of only its own language,
whereas there was little-to-no improvement for English likely due to the 
relatively large amount of training data.





\subsection{Subtask 1 - Document Classification}

\noindent\fbox{%
    \parbox{\linewidth}{%
Does a news article contain information about a past (or ongoing) event? (Binary classification)
    }%
}

\subsubsection{Experimental Approach}
The document classification subtask had the unique challenge of testing on Hindi
- a language not present in the training data. 
Therefore, we aimed to create a classifier that would perform 
the classification task across seen and unseen languages.

Similar to Subtask~2, 
we achieved this by using pre-trained multilingual embedding models that have
proven capabilities in using semantic similarity across languages. 
On top of these models, we then trained a classifier
capable of performing on other languages due to consistent embeddings.

Time constraints prevented the training of larger models 
such as XML-RoBERTa-large, 
which we believe could have lead to better results 
(based on our experience in other work).

\subsubsection{Model and Data Architecture}
Our final model was a 4-layer MLP classifier 
on top of 768-dimensional LaBSE embeddings,
trained and validated on a dataset that directly combined all 3 languages 
in the training set (split $80/20$ as internal training and validation sets). 


\subsubsection{Experimental Results}
\label{sec:org067af09}
Due to time constraints, we were unable to perform any ablation tests on the Subtask~1 data. 
Thus, we assumed that training with all languages (as in Subtask~2) 
would yield good performance and may generalize better to unseen languages. 
A single model was used for the final submission, and the results are given in Table \ref{table:subtask1}.

\begin{table}[h]
\centering 

\begin{tabular}{lrrrr}
    \hline
    Dataset & [en] & [es] & [pt] & Hindi\\
    \hline
    Val. & 0.7060 & 0.5710 & 0.6510 & -\\
    Comp. & 0.7758 & 0.6984 & 0.8121 & 0.5955\\
    Placing & 6/10 & 3/8 & 3/8 & 5/7\\
    \hline
\end{tabular}
\caption{Averaged Model Performance for Subtask~1} 
\label{table:subtask1} 

\end{table}


\subsubsection{Subtask~1 Discussion}
Performance on Spanish and Portuguese 
showed the benefits of 
training on all data instead of only individual languages. 
%
%
For the unseen language Hindi, 
it is possible that the model over-fitted to the provided languages during training - 
though it is impressive that the simple technique used is capable of domain-transfer
`out-of-the-box'.





\subsection{Subtask 3 - Event Coreference Identification}

\noindent\fbox{%
    \parbox{\linewidth}{%
Which event sentences (from Subtask 2) are about the same event? (All-vs-all linking)
    }%
}

\subsubsection{Experimental Approach}

Subtask~3 had significantly less training data 
for Spanish (11 documents) and Portuguese (21 documents) 
compared to English (596 documents) (collectively, ``ACL-St3'').
To take advantage of the larger quantity of English data, 
we made Spanish and Portuguese translations of the English training portion 
to investigate whether models improved in performance when trained on translations.

Additionally, we used an external English dataset \cite{choubey-huang-2021-automatic} 
to obtain a balanced set of $8,030$ coreferential and non-coreferential sentence pairs (``EACL-2021'') 
to investigate whether the models improve when also trained on more data.

Our final architecture was a two-stage process where we 
(i) first predict whether each sentence pair in a document is co-referential (binary classification), 
followed by 
(ii) a greedy clustering of sentences predicted to be co-referential.

For the first stage, we made use of a pre-trained m-BERT
fine-tuned as a sentence pair coreference classifier
(this returned a confidence score that any two given sentences are coreferential). 
The second stage formed clusters based upon whether 
the coreference classification estimate exceeded $0.5$, 
greedily expanding the clusters in the process.

The training data was prepared by extracting unique sentence pairs from each document, 
labelling only sentence pairs in the same cluster 
as ``coreferential'' and the others as ``non-coreferential''.

\subsubsection{Model and Data Architecture}

Our best-performing solution comprised training 
m-BERT model trained once for each individual target language, 
fine-tuned as a sentence-pair coreference classifier 
(maximising $F_{0.6}$ on the validation set when 
trained/validated on a $90/10$ split of each specific dataset).

The individual language datasets were treated separately (and these combinations were found to give the best performance):
\begin{itemize}
    \item English: ACL-St3 and EACL-2021 combined
    \item Portuguese / Spanish: For each language, we combined their respective portion of the ACL-St3 dataset, 
    and translations of the English ACL-St3 dataset into that language 
    (using output from the Google translate API, un-modified).
\end{itemize}

\subsubsection{Experimental Results}



The performance of the English language model was
marginally better 
(an uplift of around 1\% in absolute score)
when the model was trained with EACL-2021 than without
.

We found better model performance on Spanish and Portuguese 
when models were trained on Spanish and Portuguese translations of the English training data than without.

The results of our best-performing model for each language, 
scored using CoNLL-2012 average \cite{pradhan-etal-2014-scoring},
are given in Table \ref{table:subtask3}.

\begin{table}[h]
\centering
\begin{tabular}{lrrr}
\hline
Dataset & [en] & [es] & [pt]\\
\hline
Validation & 0.8990 & 0.9330 & 0.8220\\
Competition & 0.7901 & 0.8195 & 0.9061\\
Final Placing & 4/6 & 4/5 & 4/5\\
\hline
\end{tabular}
\caption{Averaged Model Performance for Subtask~3} 
\label{table:subtask3}
\end{table}

\pagebreak

\subsection{Subtask 4 - Event Extraction}

\noindent\fbox{%
    \parbox{\linewidth}{%
For a given event sentence, what is the event trigger and its arguments? (BIO sentence annotation)
    }%
}

\subsubsection{Experimental Approach}

For Subtask~4, we use a pre-trained XLM-RoBERTa
with a Token Classification head and fine-tuned it on GLOCON dataset.

As stated in our overall approach, 
we aimed to maximise our multilingual capabilities while not requiring labour intensive data collection for each new language. 
To that end, 
we make a distinction between our primary language (English) which we expect to have more data for, 
and our secondary languages (Spanish, Portuguese) 
where there is less data.
Our goal is to be able to add new secondary language capabilities 
with as little data requirements as possible. 

Following \citet{xie2018neural} and \citet{wu2020unitrans},
we apply techniques from \citet{lample2017unsupervised} 
to translate our primary language training data word-by-word into our secondary languages, 
and directly copy the entity label of each primary language word to its corresponding translated word. 
Using embeddings from \citet{bojanowski2017enriching}, 
we learn a mapping, using the MUSE library, 
from the primary to the secondary language 
making use of identical character strings between the two languages. 
To produce the word-to-word translations, 
we use the learned mapping to map 
the primary language word into the secondary language embedding space, 
and find its nearest neighbour as the corresponding translated word. 
Additionally, as described in \citet{conneau2018word}, 
we mitigated the ``hubness'' problem by using cross-domain similarity local scaling (CSLS) 
to measure the distance between the mapped embedding vector 
of the primary language word and the embedding vector of a secondary language word.  
For an illustrative example please see Tables 
\ref{table:parallel-translations-sentence} and \ref{table:parallel-translations-words}.

\begin{table}[t]
\begin{tabular}{ p{0.3cm} l } 
    \hline
& \\[\dimexpr-\normalbaselineskip+2pt]
$[en]$ & KSRTC buses were attacked at ten places. \\
$[pt]$ & Os ônibus KSRTC foram atacados em dez \\
 & lugares. \\
$[es]$ & Los autobuses KSRTC fueron atacados en \\
 & diez lugares. \\
    \hline
\end{tabular}

\caption{
\label{table:parallel-translations-sentence}
Sentence-wise translations 
(contrast with words/grammar of Table \ref{table:parallel-translations-words})
}

    \vspace{20pt}

\centering
\begin{tabular}{ l l l } 
    \hline
    $base[en]$ & $[en]\rightarrow[pt]$ & $[en]\rightarrow[es]$ \\
    \hline 
&& \\[\dimexpr-\normalbaselineskip+2pt]    
    KSRTC     & DERSA  & BIZKAIBUS \\
    buses     & ônibus & autobuses \\
    were      & foram  & fueron \\
    attacked  & atacou & atacado \\
    at        & na     & en \\
    ten       & dez     & diez \\
    places    & lugares & lugares \\
    .         & .         & . \\
    \hline
\end{tabular}
\caption{
\label{table:parallel-translations-words}
Word-by-Word translation example, allowing for consistent BIO tagging
}
\end{table}

Thus, we are able to train our model on new secondary languages 
without requiring task-specific secondary language data, 
but rather secondary language embeddings 
and bilingual primary-secondary dictionaries to create the mapping. 
For each language, our training sets consisted of 90\% of the English training data 
and the translated secondary language data, 
%
%
and our validation set was the (entire) original secondary language training data set, 
plus the remaining 10\% of the English training data%
\footnote{One dataset-specific issue : 
Care had to be taken to avoid translating the English validation set 
as it resulted in the model having access to a form of the validation set (data leakage).%
}%
.

The final classification is decoded using the Viterbi Algorithm \cite{1054010}. 
Instead of training transition probabilities based upon our limited training data, 
we instead explicitly encoded constraints (by setting selected transition probabilities to zero)
to ensure that we do not violate the BIO tagging scheme
.

\subsubsection{Experimental Results}

The results of our model for each language, are given in Table \ref{table:subtask4}.
%
There was no performance degradation between training the model on 
\{1 primary + 1 secondary language\}
vs 
\{1 primary + 2 secondary languages\}, 
which is promising for application to other secondary languages in the future.

\begin{table}[h]
\centering
\begin{tabular}{l   r r r } 
 \hline
 Dataset & [en] & [es] & [pt] \\
 \hline 
 Validation  & 82.53 & 62.17 & 72.75 \\
 Competition & 73.53 & 62.21 & 68.15 \\
 Final Placing & 2/5 & 2/4 & 2/4 \\
\hline
\end{tabular}
\caption{$F_1$ Model Performance for Subtask~4} 
\label{table:subtask4}
\end{table}

\begin{table*}[t]
\centering
\begin{tabular}{|l  | c | c | r r r r |} 
 \hline
 Model & Viterbi & W-to-W & English $F_1$ & Spanish $F_1$ & Portuguese $F_1$ & Average $F_1$ \\
 \hline 
 Baseline BERT &&& 71.54 & - & - & - \\
 MultiLingual BERT &&& 70.99 & 54.94 & 64.96 & 63.63 \\
 XLM-RoBERTa &&& 70.81 & 53.46 & 68.14 & 64.14 \\
 XLM-RoBERTa & \checkmark && 72.80 & 54.65 & 70.46 & 65.97 \\
 XLM-RoBERTa & \checkmark & \checkmark & 82.53 & 62.17 & 72.75 & 72.48 \\
\hline
\end{tabular}
\caption{Model ablation for Subtask~4 on validation set.  
`Viterbi': The BIO tagging is cleaned using Viterbi decoding.
`W-to-W': Models are trained with word-to-word translated data.}
\label{table:4-val}
\end{table*}

It is interesting to observe that
the difference in scores between validation and test sets was approximately 5\%. 
This might indicate that either that the test set has a rather different distribution 
from the validation set or that we may have biased the validation set in some manner. 

We also observe in Table \ref{table:4-val} that adding translated secondary language data 
helped to improve the performance on our \textit{primary} data. 
While we did not dig deeper into the cause, we did notice that with the translated data the model took about twice the number of epochs to converge.

\section{Discussion}

In Subtasks 1, 2 and 3, we found that our Competition performance was generally higher 
than that obtained on our own validation split of the training data.  
This surprising outcome is difficult to explain, though may be because:

\begin{itemize}
    \item Low data effects : Our validation data sets were necessarily quite small, 
    and we may have simply had a non-representative selection of harder examples in those 
    subsets
    \item Test data is `constructed' : Perhaps there are some additional statistical 
    effects that the Shared Task organisers want to analyse, and thus the test data
    distribution is intentionally different (eg: split into `easy' and `hard' subsets)
    from the training data
\end{itemize}

\section{Conclusions}

%
%

We showed that it is possible to 
achieve strong performance 
on new languages without task specific training data in the new language, 
%
provided that there is good enough training data in another language 
(English in this case) to supplement the training process.

This multilingual use-case is of commercial interest within our organisation
and we thank the organisers of the Shared Task for the opportunity 
to explore these issues using curated datasets.


\section*{Acknowledgments}


We would like to thank Ethan Phan, 
Yuan Lik Xun, 
Rainer Berger 
and Charles Poon
for their valuable input during the Shared Task, 
as well as
Handshakes 
for being supportive of this research project.

\bibliographystyle{acl_natbib}
\bibliography{acl2021}

\end{document}